\documentclass[runningheads]{llncs}

\usepackage[T1]{fontenc}
\usepackage{graphicx}
\usepackage{booktabs}
\usepackage{multirow}
\usepackage{amsmath,amssymb}
\usepackage{subcaption}
\usepackage{color}
\usepackage{hyperref}

\begin{document}

\title{HiRes: A Hierarchical Cascaded Method for Resistor Value Identification}

\titlerunning{HiRes: A Hierarchical Cascaded Method for Resistor Value Identification}
\author{Rama Y. AlHamidi\inst{1}\orcidID{0009-0004-2436-2986} \and
Aseel A. Mohamed\inst{1}\orcidID{0009-0007-5027-7963} \and
Mustafa A. Eltayeb\inst{1}\orcidID{0009-0005-8947-5567} \and
Osama Hasoneh\inst{1}\orcidID{0009-0000-5543-9377}\and
Mohammad Shaqfeh\inst{2}\orcidID{0000-0002-9330-5924}}

\authorrunning{R.\,Y. AlHamidi et al.}

\institute{Electrical \& Computer Engineering, Texas A\&M University at Qatar,
Doha, Qatar\\
\email{\{ramaalhamidi,aseelanass, m2907\}@tamu.edu} ,
\email{osama.hasoneh@qatar.tamu.edu}
\and
College of Science and Engineering, Hamad Bin Khalifa University,
Doha, Qatar\\
\email{moshaqfeh@hbku.edu.qa}}

\maketitle

\begin{abstract}
Accurate identification of resistor values from unconstrained images remains a challenging computer vision task due to variations in lighting, orientation, scale, and background complexity. This paper presents HiRes, a hierarchical cascaded pipeline for end-to-end resistor value identification directly from full-frame images. The approach combines object detection (YOLOv8n), semantic segmentation (UNet++ with EfficientNet-B2), and structured geometric decoding via projection along the resistor axis. To improve robustness, we incorporate geometric filtering, gap-preserving band separation, and validation against the E24 resistor series. Experiments across diverse real-world images show that HiRes achieves a detection mAP\textsubscript{50} of 0.9906, a segmentation mIoU of 0.8444, and an end-to-end identification accuracy of 85.8\% (95\% CI: 78.0--91.9\%), outperforming the publicly available classical baseline, CVResist, which fails to generalize beyond controlled conditions. In addition, our architecture outperforms state of the art MLLMs on our challenging test set, offering a lower cost, high efficiency, and an interpretable alternative method. These results demonstrate the effectiveness of integrating learned visual representations with structured reasoning for robust resistor interpretation. Code and dataset are available at \url{https://github.com/HiRes491/HiRes}.

\end{abstract}

\section{Introduction}
\label{sec:intro}

Resistor color codes encode resistance values through colored bands, making accurate interpretation necessary for reliable component identification during circuit analysis and prototyping. While decoding is straightforward for humans under ideal conditions, automating this process from images remains challenging due to variations in lighting, orientation, scale, and background complexity. In real-world scenarios, resistors may appear rotated, partially occluded, or embedded in cluttered scenes, making reliable band detection and interpretation difficult.

Traditional computer vision approaches address this problem using heuristic pipelines based on color thresholding, edge detection, and geometric projections~\cite{chen2016resistor}. Although effective under controlled conditions, these methods rely on handcrafted rules and fixed thresholds that do not generalize across imaging environments, suffering from band merging, fragmentation, and incorrect ordering.

Recent advances in deep learning, particularly in object detection and semantic segmentation, offer a promising alternative by learning robust visual representations directly from data. However, resistor interpretation requires both precise localization of thin, closely spaced color bands and structured reasoning to interpret their order and map them to valid resistance values. We argue that resistor value identification is best formulated not as a single vision task, but as a sequence of interdependent stages that mirror human reasoning: first localizing the resistor within a full-frame image, then segmenting its color bands at the pixel level, and finally ordering and decoding the bands into a valid resistance value according to standardized color code tables.

To address these challenges, we propose \textbf{HiRes}, a \underline{\textbf{hi}}erarchical cascaded pipeline for \underline{\textbf{res}}istor value identification that integrates learned visual features with structured reasoning for end-to-end identification from unconstrained images. The architecture is hierarchical in that each stage operates at a progressively finer spatial granularity. At the scene level, a YOLOv8n object detector localizes resistor instances and produces cropped region(s) of interest. At the pixel level, each crop is segmented using UNet++ with an EfficientNet-B2 encoder. At the band level, the segmentation mask is projected along the resistor's principal axis to extract an ordered band sequence. The decoded resistance value is validated against the IEC~60062 standard~\cite{iec60062} and the E24 preferred value series~\cite{iec60063}.

The main contributions of this work are as follows: (1) A hierarchical cascaded pipeline that combines learned detection and band segmentation with structured geometric decoding for complete unconstrained resistor value identification. (2) A gap-preserving band extraction mechanism that operates on raw projection vote counts to prevent Gaussian smoothing from merging adjacent same-color bands. (3) A comparative evaluation against general-purpose vision-language models, showing that current MLLMs struggle with precise color-band ordering and symbolic resistor-code decoding under unconstrained imaging conditions.

\section{Related Work}
\label{sec:related_work}

Prior work on resistor value identification has primarily relied on
classical image processing pipelines with limited adoption of learned
classifiers.

Chen and Wang~\cite{reading2015resistor,chen2016resistor} developed a
full system for orientation detection, body segmentation, band extraction,
and value decoding, achieving over 90\% accuracy but requiring a custom
hardware setup with controlled lighting, limiting applicability outside
laboratory settings.
Demir et al.~\cite{demir2018resistor} proposed a real-time mobile
implementation using bilateral filtering and color matching in HSV space,
though the method requires manual alignment to an on-screen guide and
provides no automatic detection or segmentation.
Serban and Hobincu~\cite{serban2021resistor} used adaptive thresholding
and peak detection, evaluated on 12 resistors against a white background
with an industrial camera.
Wibawanto et al.~\cite{wibawanto2023resistor} applied ant colony
optimization for combinatorial decoding, achieving 85\% on 20 images but
requiring over 40 seconds and no support for unconstrained scenes.

A smaller body of work has incorporated learned classifiers.
Liu et al.~\cite{liu2020resistor} trained an encoder-decoder network to
segment color-ring resistors from PCB images, but targeted on-board
inspection rather than standalone value reading and did not perform
color-code decoding.
Li et al.~\cite{li2017resistor} combined Retinex-based illumination
normalization with a neural-network color classifier using a monochrome
industrial camera.
Muminovic and Sokic~\cite{sokic2019resistor} classified band colors via
SVMs on RGB descriptors, evaluated on resistors photographed against a
uniform white background.
None of these addresses the full pipeline from unconstrained input through
band-level segmentation to value decoding; all operate under constrained
conditions or require prior isolation of the resistor.

We provide, to our knowledge, the first publicly released unconstrained full-frame evaluation set with code, trained weights, and end-to-end value-level metrics for resistor value identification. 

\section{Dataset}
\label{sec:datasets}

Table~\ref{tab:datasets} summarizes the datasets used for detection and segmentation. Detection data were sourced from Roboflow; segmentation data were annotated using the Datature platform.

\subsection{Detection Dataset}

\begin{table}[t]
  \caption{Dataset Summary}
  \label{tab:datasets}
  \centering
  \renewcommand{\arraystretch}{1.15}
  \begin{tabular}{llccc}
    \toprule
    \textbf{Task} & \textbf{Source} & \textbf{Train} & \textbf{Val} & \textbf{Test} \\
    \midrule
    Detection & Roboflow & 2256 & 564 & 180 \\
    \midrule
    Segmentation & Datature & 660 & 140 & 106 \\
    \bottomrule
  \end{tabular}
\end{table}

The detection stage was trained on 3000 images collected from multiple public Roboflow datasets~\cite{resistorband_dataset,resistors7wdzf_dataset,resistorsdooos_dataset,dpaml2024rc} and supplemented with photographs captured by the project team. The dataset was formatted for single-class object detection (\textit{resistor}), with images of ceramic capacitors~\cite{ceramiccaps_dataset} included as hard negatives to encourage the detector to learn resistor-specific features rather than relying on coarse shape cues alone. This change was added after initial tests on photos with diverse components showed the object detector significantly confused ceramic capacitors and resistors more than other classes. The final split consisted of 2256 training, 564 validation, and 180 test images.

\subsection{Segmentation Dataset}

\begin{table}[t]
  \caption{Segmentation Color Class Label Map
  (IEC 60062~\cite{iec60062} / 60063~\cite{iec60063})}
  \label{tab:colormap}
  \centering
  \renewcommand{\arraystretch}{1.15}
  \begin{tabular}{clcc}
    \toprule
    \textbf{ID} & \textbf{Color} & \textbf{Hex Code} & \textbf{Digit / Multiplier / Role} \\
    \midrule
    0  & Background & \texttt{\#FFCDD2} & --- \\
    1  & Black      & \texttt{\#000000} & 0 / $\times 1$ \\
    2  & Blue       & \texttt{\#004AAD} & 6 / $\times 10^{6}$ \\
    3  & Brown      & \texttt{\#5E3831} & 1 / $\times 10$ \\
    4  & Gold       & \texttt{\#EBBF7C} & $\times 10^{-1}$ / $\pm 5\%$ tolerance \\
    5  & Green      & \texttt{\#00BF63} & 5 / $\times 10^{5}$ \\
    6  & Grey       & \texttt{\#585A59} & 8 / $\times 10^{8}$ \\
    7  & Orange     & \texttt{\#FF914D} & 3 / $\times 10^{3}$ \\
    8  & Violet     & \texttt{\#5E17EB} & 7 / $\times 10^{7}$ \\
    9  & Red        & \texttt{\#E43232} & 2 / $\times 10^{2}$ \\
    10 & Silver     & \texttt{\#CDCDCD} & $\times 10^{-2}$ / $\pm 10\%$ tolerance \\
    11 & White      & \texttt{\#FFFFFF} & 9 / $\times 10^{9}$ \\
    12 & Yellow     & \texttt{\#F4CB24} & 4 / $\times 10^{4}$ \\
    \bottomrule
  \end{tabular}
\end{table}

The segmentation model was trained on 800 images~\cite{resistor_value_training_Computer_Vision_Model} of 4-band and 5-band resistors, annotated at the pixel level. All images were cropped prior to annotation to match the input distribution encountered during inference at that stage. The dataset was split into 660 training and 140 validation images. A separate set of 106 images, disjoint from both splits, is used for end-to-end evaluation with value-level ground truth only (Section~\ref{sec:results_e2e}). The class taxonomy comprises 13 classes corresponding to the standard resistor color code (Table~\ref{tab:colormap}). Gold and silver are treated as tolerance bands rather than significant digits or multipliers in the decoding stage.

\section{Methodology}
\label{sec:method}

\subsection{Pipeline Overview}

\begin{figure}[t]
    \centering
    \includegraphics[width=0.95\textwidth]{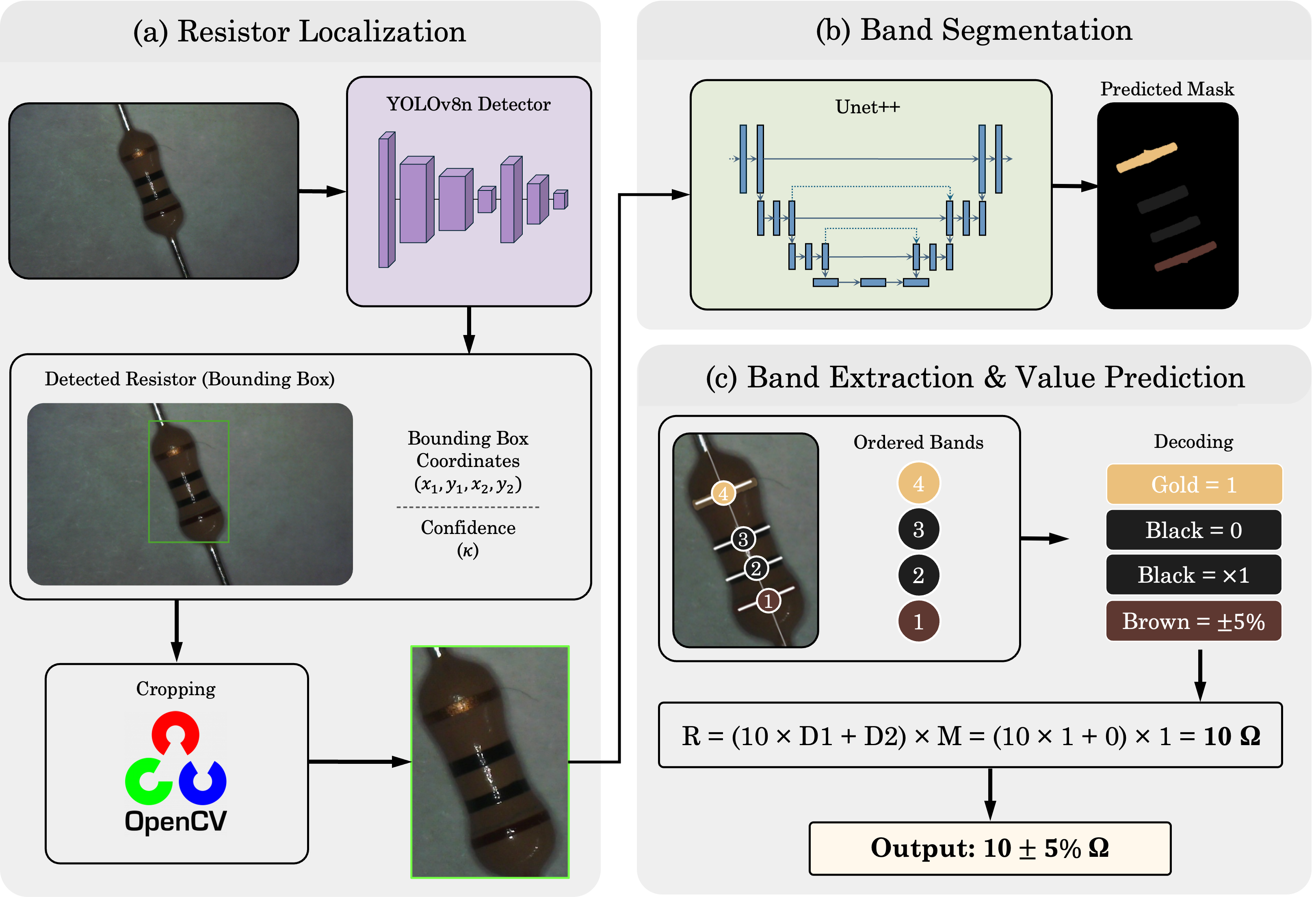}
    \caption{Overview of the proposed HiRes pipeline. (a) YOLOv8n detects and crops the resistor from the input image. (b) UNet++ performs pixel-wise segmentation of color bands. (c) Band extraction and ordering are carried out along the resistor axis, followed by decoding to obtain the final resistance value.}
    \label{fig:overview}
\end{figure}

Given an input image containing one or more resistors, the objective is to produce a structured resistance value, including both magnitude and tolerance, without manual intervention. The pipeline decomposes this into three sequential stages (Fig.~\ref{fig:overview}), each operating at progressively finer spatial granularity.

First, a YOLOv8n object detector localizes resistor instances and extracts cropped regions of interest. Second, each crop is processed by a two-pass UNet++ segmentation module that assigns a color class label to every pixel, producing a refined band-level mask. Third, a band extraction stage projects the segmentation output along the resistor's principal axis to identify individual bands, determine reading direction, and generate an ordered color sequence.

\subsection{Object Detection}

The first stage of the pipeline localizes resistor instances within the input image by fine-tuning a YOLOv8n detector~\cite{jocher2023ultralytics} to map full-frame images to candidate bounding boxes. YOLOv8n (the nano variant of the YOLOv8 family) is selected due to its favorable trade-off between accuracy, inference speed, and model size~\cite{aboyomi2023yolo}.

During inference, the detection confidence threshold is set to 0.01 to maximize recall, ensuring that no resistor instance is discarded prematurely; false positives are filtered by the segmentation and decoding stages, which reject crops that do not yield a valid band sequence. Each detection produces an axis-aligned bounding box $(x_1,y_1,x_2,y_2)$ along with an associated confidence score $c_{det}\in[0,1]$.

Each predicted bounding box is expanded by 10\% on each side and cropped. The resulting regions of interest (ROIs) are resized and padded to $512\times512$ pixels while preserving aspect ratio, ensuring consistent input dimensions for segmentation and preserving contextual information at the ends of the resistor where color bands may otherwise be truncated.

\subsection{Band Segmentation}

\subsubsection{Model Architecture.}
The segmentation stage employs a UNet++ architecture~\cite{zhou2018unetplusplus} with an EfficientNet-B2 encoder~\cite{tan2019efficientnet} to classify each pixel into one of 13 classes corresponding to the resistor color code.

UNet++ is chosen over alternatives such as DeepLabV3+ because its nested, dense skip connections reduce the semantic gap between encoder and decoder feature maps, enabling more precise boundary localization for thin and closely spaced bands where dilated-convolution architectures tend to produce coarser predictions. EfficientNet-B2 is selected as the encoder for its compound-scaled balance of depth, width, and resolution: B0 lacks the capacity for reliable 13-class discrimination at band boundaries, while larger variants (B4+) risk overfitting given only 660 training images. B2 pretrained on ImageNet provides sufficient expressiveness while generalizing effectively under varying lighting, viewpoints, and image quality. Input dimensions match the detection stage ($512 \times 512$, aspect-ratio-preserving padding).

\subsubsection{Training Strategy.}
The model is trained using a composite loss function that combines Dice loss~\cite{milletari2016vnet} and Focal loss~\cite{lin2017focal}.

\begin{equation}
    L=0.5\cdot  \mathrm{DiceLoss}+0.5\cdot \mathrm{FocalLoss}(\gamma=2.0)
\end{equation}

Dice loss maximizes overlap between predicted and ground-truth regions, which is effective for thin structures such as resistor bands but does not explicitly address class imbalance. Focal loss ($\gamma=2.0$) complements it by down-weighting well-classified background pixels and focusing gradient updates on challenging regions such as band boundaries and visually similar classes (e.g., brown vs.\ red).

Optimization is performed using the Adam optimizer with a learning rate of $2 \times 10^{-4}$ and weight decay of $1 \times 10^{-4}$. A cosine annealing learning rate scheduler is employed with $T_{\max} =150$ and a minimum learning rate of $1 \times 10^{-6}$, enabling gradual refinement of model weights during training. To address class imbalance in the dataset, a WeightedRandomSampler is used during training. Each image is assigned a sampling probability proportional to the inverse frequency of its rarest class, ensuring that samples containing underrepresented color bands are more frequently selected. In addition, a scale normalization strategy is applied at data loading time. Each training image is cropped to the bounding box of its segmentation mask with an additional 20\% padding and then resized and padded to $512 \times 512$. This normalization reduces variation in resistor scale and ensures consistent representation of band widths across the dataset.

\subsubsection{Data Augmentation.}
Data augmentation is applied during training to improve robustness to variations in orientation, illumination, and imaging conditions. Geometric transformations include horizontal and vertical flipping, random rotations up to $\pm 45^\circ$, and affine transformations with moderate scaling and translation. Photometric transformations include brightness and contrast adjustment, gamma correction, Gaussian noise, Gaussian blur, CLAHE, and coarse dropout for occlusion robustness. Hue and saturation perturbations are intentionally constrained to small ranges (hue shifts within $\pm 5^\circ$) to prevent alteration of class identity, as color is the primary discriminative feature in resistor band classification.

\subsubsection{Coarse-to-Fine Refinement.}
\begin{figure}[t]
    \centering
    \includegraphics[width=0.95\textwidth]{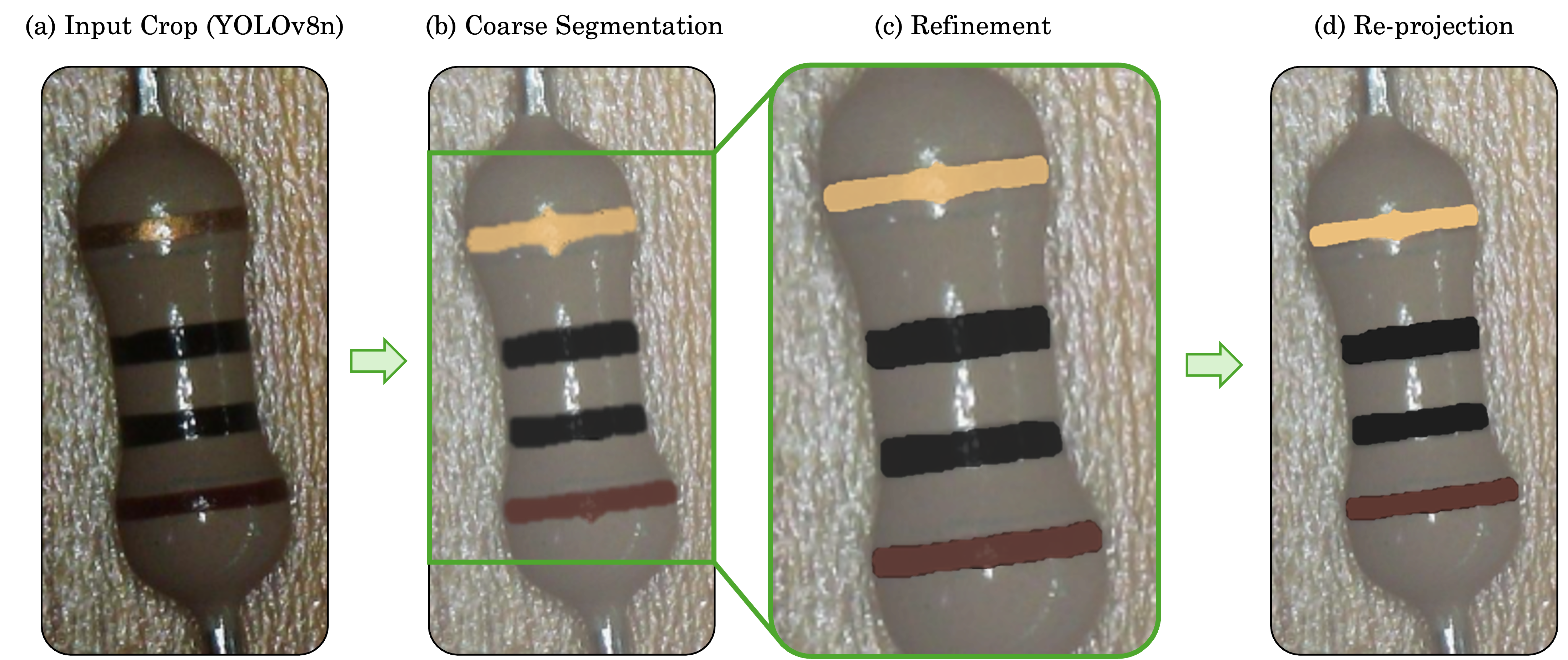}
    \caption{Coarse-to-fine segmentation strategy in HiRes. (a) YOLOv8n crop of the detected resistor. (b) Initial coarse segmentation highlighting candidate band regions, with the selected crop for refinement. (c) High-resolution refinement improves boundary accuracy and continuity of thin bands. (d) Refined predictions are re-projected onto the original image for final band localization.}
    \label{fig:ctf}
\end{figure}

A two-pass refinement strategy is applied to each detected crop, as illustrated in Fig.~\ref{fig:ctf}. In the first pass, the full crop is resized and padded to $512\times512$, normalized using ImageNet statistics, and passed through the segmentation model to produce a coarse mask. In the second pass, the non-background region of the coarse mask is cropped with 20\% padding and reprocessed by the same model; the refined prediction is then mapped back to the original coordinates, concentrating model capacity on the resistor body and improving localization of narrow, low-contrast bands.

\subsection{Band Extraction}

The band extraction stage converts the pixel-level segmentation mask $M \in \{0,\ldots,12\}^{H \times W}$ into an ordered sequence of discrete color bands. The resistor's principal axis is estimated from all non-background pixel positions by applying PCA to the point cloud $\{(x_i, y_i) : M(y_i,x_i)>0\}$. The $2\times2$ sample covariance matrix $\mathbf{\Sigma}=\frac{1}{n-1}\mathbf{X}^{\top}\mathbf{X}$, where $\mathbf{X}$ is the mean-centered coordinate matrix, is decomposed via eigenanalysis, and the eigenvector with the largest eigenvalue defines the unit axis vector $\hat{\mathbf{a}}$ with origin $\mathbf{o}$ at the point cloud centroid. Each pixel is projected to obtain a scalar position along the resistor:
\begin{equation}
    t_i = (\mathbf{p}_i - \mathbf{o}) \cdot \hat{\mathbf{a}}
\end{equation}

The projection range is discretized into $N = 200$ uniform bins and pixel votes are accumulated per class into a histogram $V \in \mathbb{R}^{N \times 13}$. Each class curve is Gaussian-smoothed ($\sigma = 2.5$ bins) to fill small intra-band gaps arising from segmentation noise, and the dominant class per bin is assigned as:
\begin{equation}
    \hat{c}[b] = \arg\max_{c \geq 1}\; \tilde{V}[b, c]
\end{equation}
where $\tilde{V}$ denotes the smoothed histogram. Background bins are identified using raw (unsmoothed) vote counts: a bin is marked background if $\sum_{c \geq 1} V[b,c] = 0$. This prevents Gaussian smoothing from bridging true physical gaps between adjacent bands, which would otherwise merge distinct bands and introduce systematic errors in band count and multiplier estimation.

Contiguous runs of identical dominant class in the binned signal are extracted by run-length encoding. Each run spanning at least $\delta = 4$ bins and covering at least 40 pixels is retained as a band candidate. The centroid and bounding box of each band are recovered from the original pixel coordinates whose projection falls within the run's bin range.

Detected bands are filtered by an absolute area floor and a relative threshold requiring each band to occupy at least 20\% of the median band area.

Reading direction is determined by tolerance band position. Gold or silver at one end is unambiguous; in their absence, extended tolerance colors (brown, red, green, blue, violet, grey) are used for five-band resistors. When both ends are valid, the largest inter-band gap identifies the tolerance end. Remaining ambiguity is resolved by secondary heuristics: reversing if the first band is black (invalid leading digit), or preferring the end whose terminal band is narrower. After orientation is fixed, misplaced leading tolerance bands are stripped and duplicate trailing tolerance bands are collapsed.

\subsection{Resistance Decoding}

The ordered band sequence is decoded according to the IEC 60062 resistor color code standard~\cite{iec60062}. Sequences with fewer than three bands are rejected, and sequences with more than five are reduced to the five largest by area before decoding. Each band is interpreted according to its position in the sequence, where bands may represent significant digits, a multiplier, or tolerance. Three configurations are supported: 3-band (two digits, multiplier, implicit $\pm 20\%$ tolerance), 4-band (two digits, multiplier, explicit tolerance), and 5-band (three digits, multiplier, explicit tolerance).

The resistance value is computed as:
\begin{equation}
    R = \begin{cases}
    (10D_1 + D_2)\times M & \text{3- and 4-band} \\[4pt]
    (100D_1 + 10D_2 + D_3)\times M & \text{5-band}
    \end{cases}
\end{equation}

\noindent where $D_i$ are the digit values and $M$ is the multiplier obtained from the corresponding band color according to Table~\ref{tab:colormap}. The computed resistance is converted into a human-readable format with appropriate unit scaling (e.g., $\Omega$, k$\Omega$, M$\Omega$).

Tolerance values are determined from the final band in the sequence using the mapping defined in Table~\ref{tab:colormap}. For configurations where no tolerance band is present (e.g., 3-band resistors), a default tolerance of $\pm 20\%$ is assumed.

For configurations producing invalid or nonstandard values, the decoded resistance is validated against the E24 preferred resistor series~\cite{iec60063}. If no valid E24 value exists within 1\% of the prediction, the smallest detected band is removed and decoding retried; if still invalid, the second-smallest is removed instead.

\section{Results}
\label{sec:results}

\subsection{Object Detection}
\label{sec:results_yolo}

\begin{figure}[t]
    \centering
    \includegraphics[width=\textwidth]{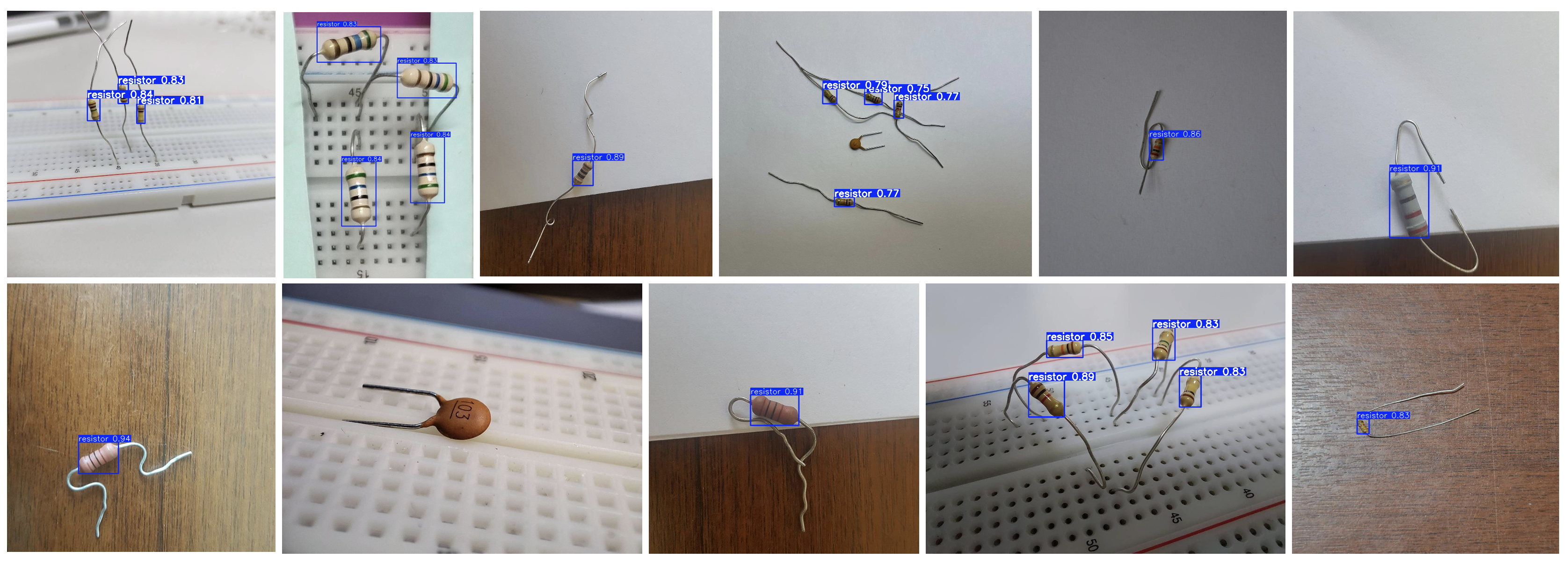}
    \caption{Qualitative results of the YOLOv8n-based resistor detector on resistor images.}
    \label{fig:yolo_results}
\end{figure}

The YOLOv8n detector achieves an F1 score of 0.9945 on a held-out test set of 180 images. Table~\ref{tab:yolo_results} summarizes precision, recall, F1-score, and mAP. On a diverse evaluation set containing varying backgrounds, lighting conditions, and scenes with multiple resistors and capacitors, the model produces one incorrect detection (179 of 180 images are correct). Representative detection results are shown in Fig.~\ref{fig:yolo_results}. The consistently high performance across variations in scale, orientation, and scene complexity indicates strong generalization beyond the training distribution.

\begin{table}[t]
  \caption{YOLOv8n Detection Results}
  \label{tab:yolo_results}
  \centering
  \renewcommand{\arraystretch}{1.15}
  \begin{tabular}{ccccc}
    \toprule
    \textbf{P} & \textbf{R}
        & \textbf{F1} & \textbf{mAP\textsubscript{50}}
        & \textbf{mAP\textsubscript{50:95}} \\
    \midrule
    0.9945 & 0.9945 & \textbf{0.9945} & 0.9906 & 0.8569 \\
    \bottomrule
  \end{tabular}
\end{table}

\subsection{Band Segmentation}
\label{sec:results_seg}

The segmentation model achieves a best validation mIoU of 0.8444 at epoch 139, with strong performance across most color classes. As shown in Table~\ref{tab:segmentation_iou}, high IoU scores are obtained for distinct colors such as blue (0.9412) and red (0.9285), while lower performance is observed for visually ambiguous classes such as grey (0.5635) and silver (0.6602). This degradation is primarily due to low contrast and similarity to neighboring colors, particularly under challenging lighting conditions.

\begin{figure}[t]
    \centering
    \includegraphics[width=0.8\textwidth]{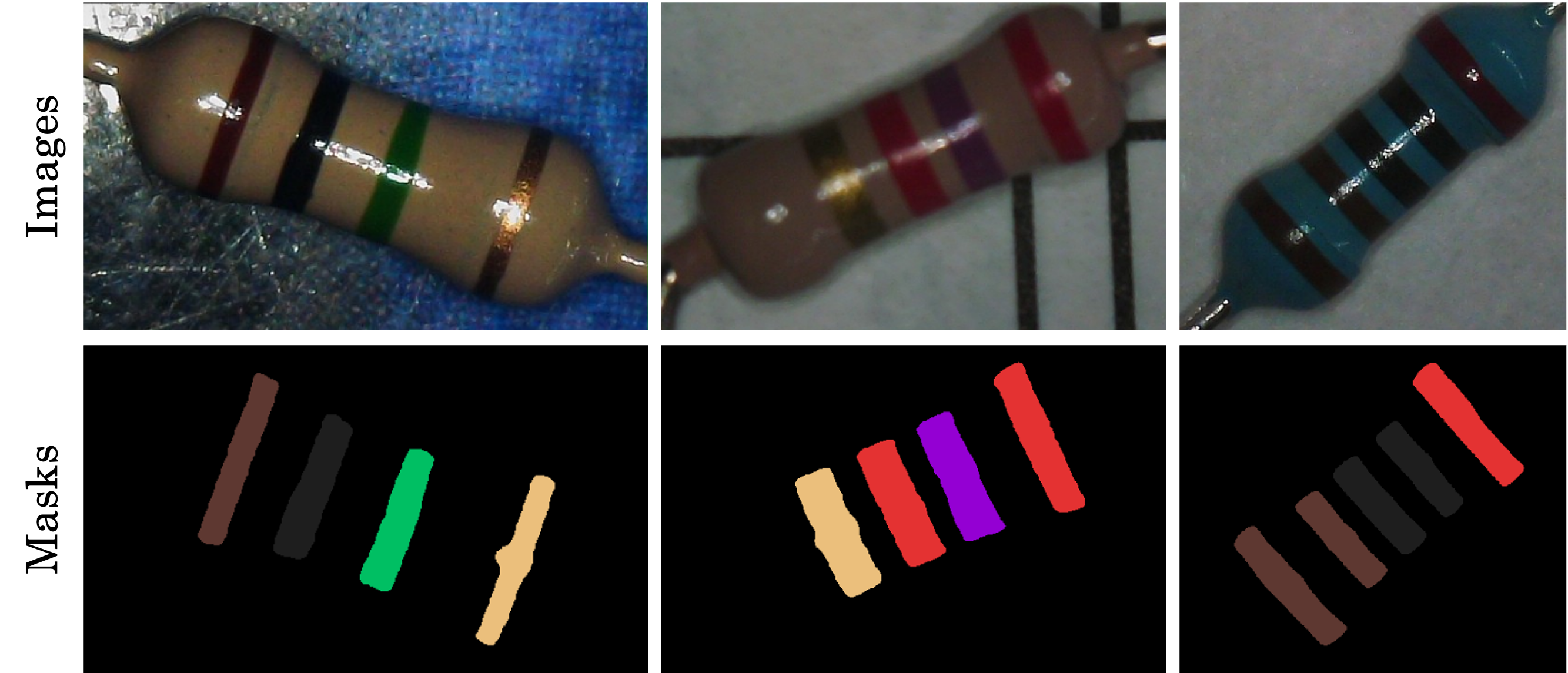}
    \caption{Qualitative segmentation results of the proposed UNet++ model.}
    \label{fig:seg_results}
\end{figure}

Qualitative results in Fig.~\ref{fig:seg_results} further demonstrate that the model accurately segments thin and closely spaced resistor bands across varying orientations, body colors, band numbers, and backgrounds. The predicted masks preserve the geometric structure of the bands, enabling reliable downstream processing. Minor errors are consistent with the quantitative results, typically occurring between similar color pairs such as brown and black.

\begin{table}[t]
  \caption{Per-class IoU}
  \label{tab:segmentation_iou}
  \centering
  \scriptsize
  \setlength{\tabcolsep}{2.5pt}
  \renewcommand{\arraystretch}{1.05}

  \begin{tabular}{lcccccccccccc}
    \toprule
    Class
    & Blk & Blu & Brn & Gld & Grn & Gry
    & Org & Red & Slv & Vio & Wht & Ylw \\
    \midrule
    IoU
    & 0.9118 & 0.9412 & 0.8964 & 0.9046 & 0.8687 & 0.5635
    & 0.8926 & 0.9285 & 0.6602 & 0.9045 & 0.8351 & 0.8257 \\
    \midrule
    Mean & \multicolumn{12}{c}{\textbf{0.8444}} \\
    \bottomrule
  \end{tabular}
\end{table}

\subsection{End-to-End Pipeline}
\label{sec:results_e2e}

\begin{figure}[t]
    \centering
    \includegraphics[width=\textwidth]{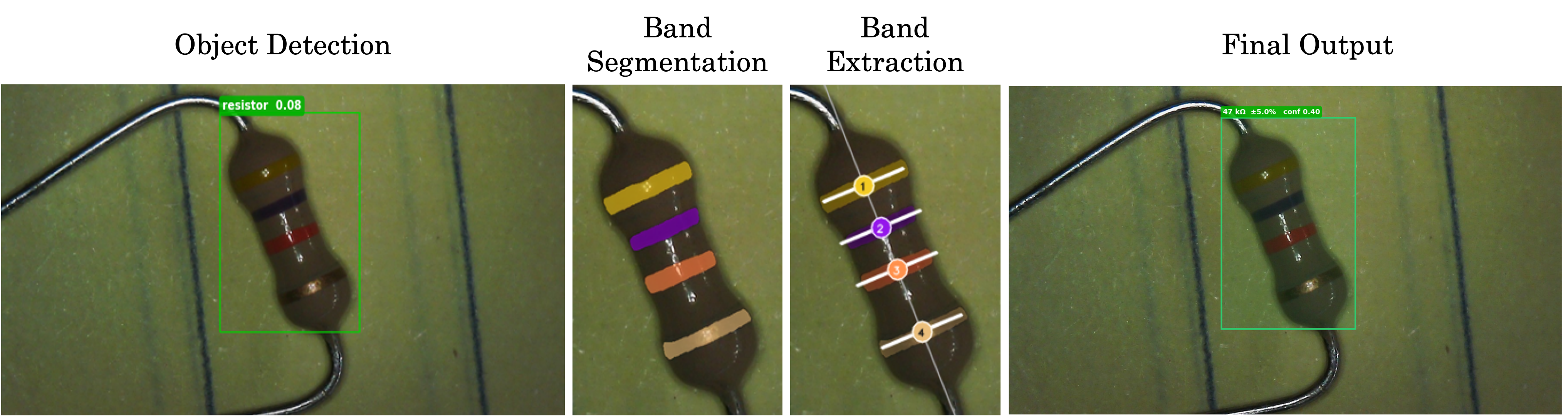}
    \caption{End-to-end pipeline demonstration of HiRes.}
    \label{fig:pipeline_example}
\end{figure}

An example of the complete pipeline in action is shown in Fig.~\ref{fig:pipeline_example}. The system processes an input image through sequential stages of detection, segmentation, band extraction, and decoding to produce the final resistance value. The full pipeline is evaluated on a test set of 106 images covering both four-band and five-band configurations, separate from the 180-image detection test set in Section~\ref{sec:results_yolo}. The YOLOv8n detector localizes 103 of 106 resistors on this set, but end-to-end accuracy depends on the combined performance of all stages. As summarized in Table~\ref{tab:accuracy comp}, the proposed method achieves an overall accuracy of 85.8\% (95\% CI: 78.0--91.9\%).

Performance differs between configurations: four-band resistors achieve higher accuracy (90.4\%) compared to five-band resistors (81.5\%), reflecting the increased difficulty of accurately segmenting and ordering additional bands. Of the 15 errors, 3 are detection failures, 6 arise from segmentation misclassification, and 6 from incorrect band ordering or decoding; 10 of the 15 involve five-band resistors. On an NVIDIA RTX 3050 Laptop GPU, the full pipeline processes a single image in 163\,ms (detection: 39\,ms, segmentation: 124\,ms, band extraction and decoding: $<$5\,ms).

\begin{table}[t]
  \caption{End-to-end pipeline comparison against general-purpose vision-language models.}
  \label{tab:accuracy comp}
  \centering
  \scriptsize
  \setlength{\tabcolsep}{3pt}
  \renewcommand{\arraystretch}{1.1}

  \begin{tabular}{lccccc}
    \toprule
    \textbf{Model}
    & \textbf{Overall}
    & \textbf{4-band}
    & \textbf{5-band}
    & \textbf{Latency}
    & \textbf{\$/sample} \\
    \midrule

    HiRes (ours)
    & \textbf{85.8\% (91/106)}
    & \textbf{90.4\% (47/52)}
    & \textbf{81.5\% (44/54)}
    & \textbf{0.163s}
    & \textbf{N/A} \\

    \midrule

    Opus 4.5
    & 21.7\% (23/106)
    & 34.6\% (18/52)
    & 9.3\% (5/54)
    & 3.90s
    & \$0.73 \\

    GPT-5.5
    & 37.7\% (40/106)
    & 48.1\% (25/52)
    & 27.8\% (15/54)
    & 18.18s
    & \$0.025 \\

    Gemini-3.1-Pro
    & 46.2\% (49/106)
    & 69.2\% (36/52)
    & 24.1\% (13/54)
    & 30.42s
    & \$0.032 \\

    Qwen3.6-plus
    & 41.5\% (44/106)
    & 59.6\% (31/52)
    & 24.1\% (13/54)
    & 179.96s
    & \$0.019 \\

    \bottomrule
  \end{tabular}
\end{table}

\subsection{Baseline Comparison}
\label{sec:baseline}

Because no standardized benchmark or widely adopted public baseline exists for unconstrained end-to-end resistor value identification, direct comparison with prior methods under matched conditions is not possible. Existing resistor-reading systems either require controlled imaging setups, manual alignment, pre-cropped inputs, or do not release source code and evaluation data. We therefore include two complementary comparisons: first, a comparison against CVResist~\cite{SupreethRao99_CVResist}, the only publicly available classical implementation we could identify; and second, a comparison against general-purpose vision-language models as task-agnostic reference points. 

\begin{figure}[t]
    \centering
    \includegraphics[width=0.85\textwidth]{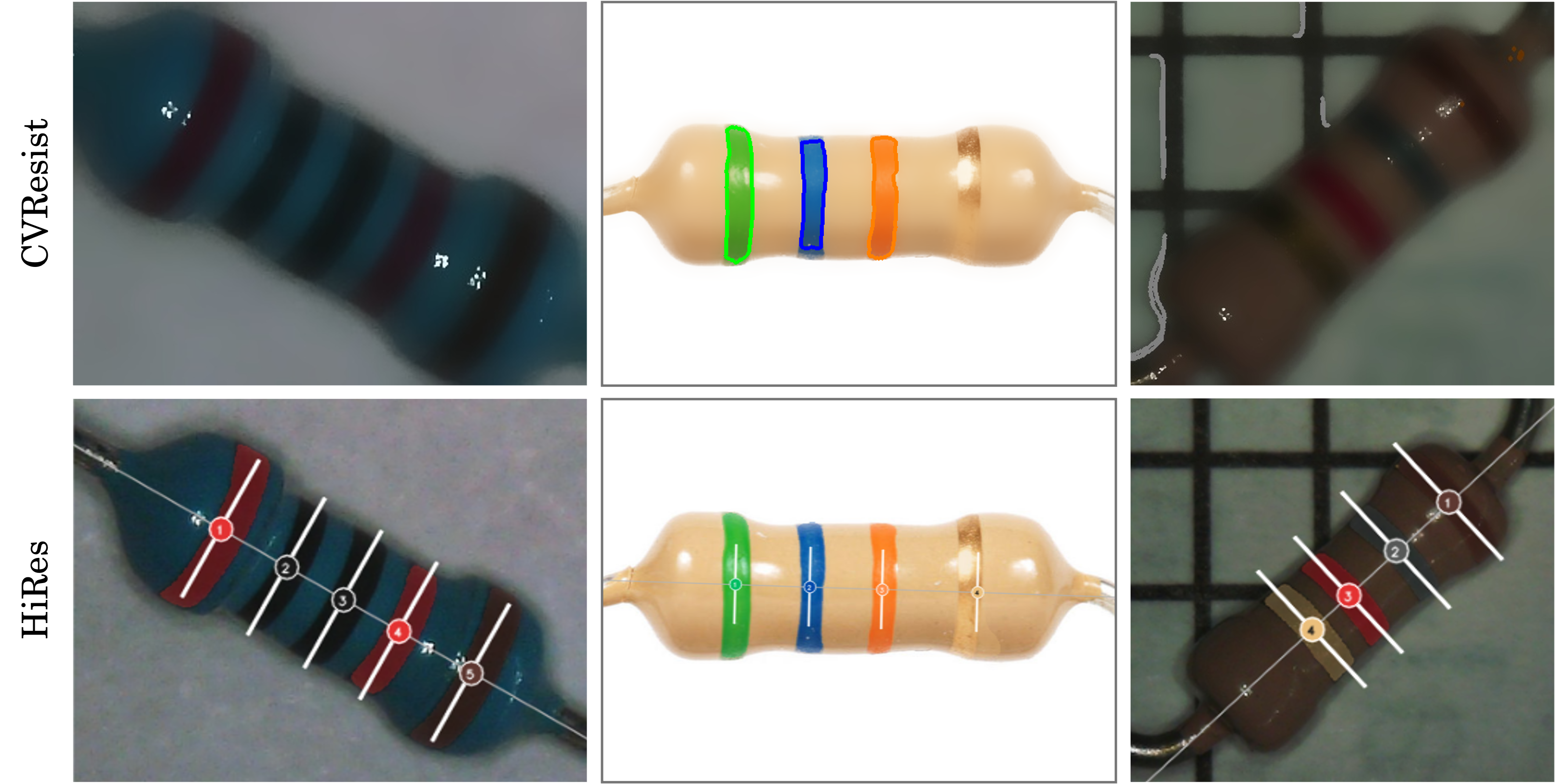}
    \caption{Qualitative comparison between the classical CVResist baseline (top row) and the proposed HiRes pipeline (bottom row). Under challenging conditions such as blur, reflections, and complex backgrounds, CVResist exhibits band merging and missed detections, while the proposed method maintains accurate band localization, ordering, and consistency.}
    \label{fig:baseline_comparison}
\end{figure}

CVResist~\cite{SupreethRao99_CVResist} operates on cropped resistor images using thresholding and edge-based segmentation in HSV space, followed by peak detection along the resistor axis and rule-based IEC decoding. To match its expected input format, we supplied CVResist with resistor crops obtained from our detection stage rather than full-frame images. Even under this favorable setting, CVResist achieved 0/106 correct identifications on our test set. This failure is primarily due to fixed HSV thresholds and edge-based band separation, which do not generalize to the illumination variation, blur, reflections, body colors, and background complexity present in the evaluation images. Fig.~\ref{fig:baseline_comparison} illustrates representative failure modes.

Table~\ref{tab:accuracy comp} also compares HiRes with several general-purpose vision-language models. These models are not domain-specific baselines; rather, they serve as task-agnostic references for evaluating whether current general visual reasoning systems can perform resistor value decoding from unconstrained images. The strongest vision-language model, Gemini-3.1-Pro~\cite{gemini31pro}, achieved 46.2\% accuracy, which is 39.6 percentage points lower than HiRes. It also required 30.42 seconds per sample, approximately 186$\times$ slower than HiRes, with an estimated API cost of \$0.032 per sample. The remaining models, GPT-5.5~\cite{GPT5.5}, Opus 4.5~\cite{opus4.5}, and Qwen3.6-Plus~\cite{qwen2026qwen36}, similarly achieved substantially lower accuracy and higher latency than the proposed task-specific pipeline.

\section{Ablation Study}
\label{sec:ablation}

\begin{table}[t]
  \caption{Ablation study of pipeline components. Each row removes one
           component from the full pipeline while keeping all others active.}
  \label{tab:ablation}
  \centering
  \renewcommand{\arraystretch}{1.15}
  \begin{tabular}{lccc}
    \toprule
    \textbf{Configuration} & \textbf{4-band} & \textbf{5-band} & \textbf{Overall} \\
    \midrule
    Full pipeline                          & 47/52 & 44/54 & \textbf{91/106 (85.8\%)} \\
    ($-$ coarse-to-fine refinement)     & 40/52 & 42/54 & 82/106 (77.4\%) \\
    ($-$ band gap detection)            & 45/52 &  9/54 & 54/106 (50.9\%) \\
    ($-$ detection stage)        & ---   & ---   &  0/106 (0.0\%)  \\
    \bottomrule
  \end{tabular}
\end{table}

Table~\ref{tab:ablation} isolates the effect of two key components introduced in this work: the coarse-to-fine segmentation refinement (Section~4.3.4) and the raw-vote band gap detection (Section~4.4). In each experiment, the targeted component is removed while the rest of the pipeline remains unchanged. Detection is included as a baseline to confirm that localization is a prerequisite; without it, the system receives full-frame images and produces no usable output.

Removing the refinement pass reduces accuracy by 8.5\%. Without it, the segmentation model operates on the full YOLO crop where the resistor may occupy a small fraction of the 512$\times$512 canvas, yielding lower effective resolution on band regions and noisier color boundaries. The drop is moderate and distributed across both resistor types.

Removing gap detection has a substantially larger effect, reducing accuracy from 85.8\% to 50.9\%. The impact falls almost entirely on 5-band resistors, which drop from 44/54 to 9/54. The failure is systematic: Gaussian smoothing in the 1D projection fills narrow physical gaps between adjacent same-color bands, merging two bands into one and shifting the decoded value by a factor of ten. Gap detection prevents this by preserving bins where no pixels were observed in the raw vote histogram, maintaining correct band separation regardless of smoothing. 4-band resistors are only marginally affected ($-$2), as they rarely exhibit adjacent same-color bands.

Removing E24 validation produced no change in end-to-end accuracy, confirming that the observed failures are band-count errors (merged or missing bands) rather than marginal decoding mismatches that E24 correction could address.

\section{Conclusion}

This paper presented HiRes, a hierarchical cascaded pipeline for
end-to-end resistor value identification from unconstrained images. By
combining YOLOv8n-based detection, UNet++ pixel-level band segmentation,
and structured geometric decoding, the system achieves 85.8\% end-to-end
accuracy (95\% CI: 78.0--91.9\%) on a 106-image test set spanning diverse
real-world conditions. Ablation studies show that gap-preserving band
separation is the single most impactful component, contributing a 34.9\% improvement and proving essential for correct five-band interpretation. A notable finding of this work is the absence of any publicly available baseline capable of operating on unconstrained images; existing methods either lack released implementations or require controlled
imaging setups, underscoring the need for standardized benchmarks in this
domain. Compared to general-purpose vision-language models, our architecture provides higher accuracy results at a fraction of the cost and latency.

Despite these results, five-band resistors remain more challenging due to
increased sensitivity to segmentation errors and band ordering ambiguity.
Future work will focus on improving robustness for five-band configurations
through enhanced separation of closely spaced bands and improved handling
of visually ambiguous colors, scaling to larger and more diverse evaluation
sets, and exploring deployment on embedded or mobile platforms for
real-time in-field component identification. Our approach enables practical applications in automated circuit inspection and manufacturing quality control, where reliable identification of resistor values from images is essential. We release our code, trained
weights, and test set to serve as a reproducible baseline and encourage
future comparisons.


\bibliographystyle{splncs04}
\bibliography{references}

@misc{resistorband_dataset,
    title = {Resistor Band Dectection Dataset},
    type = {Open Source Dataset},
    author = {Basic},
    howpublished = {\url{https://universe.roboflow.com/resistors-leuoc/resistor-band-dectection}},
    url = {https://universe.roboflow.com/resistors-leuoc/resistor-band-dectection},
    journal = {Roboflow Universe},
    publisher = {Roboflow},
    year = {2025},
    month = {apr},
    note = {visited on 2026-03-18}
}

@misc{resistorsdooos_dataset,
    title = {Resistors Dataset},
    type = {Open Source Dataset},
    author = {blaze},
    howpublished = {\url{https://universe.roboflow.com/blaze-k81pv/resistors-dooos}},
    url = {https://universe.roboflow.com/blaze-k81pv/resistors-dooos},
    journal = {Roboflow Universe},
    publisher = {Roboflow},
    year = {2025},
    month = {feb},
    note = {visited on 2026-03-18}
}

@misc{resistors7wdzf_dataset,
    title = {resistors Dataset},
    type = {Open Source Dataset},
    author = {relipski@ncsu.edu},
    howpublished = {\url{https://universe.roboflow.com/relipski-ncsu-edu/resistors-7wdzf}},
    url = {https://universe.roboflow.com/relipski-ncsu-edu/resistors-7wdzf},
    journal = {Roboflow Universe},
    publisher = {Roboflow},
    year = {2022},
    month = {apr},
    note = {visited on 2026-03-18}
}

@misc{ceramiccaps_dataset,
    title = {resistors\_and\_ceramic\_caps Dataset},
    type = {Open Source Dataset},
    author = {{491 Resistor Classification}},
    howpublished = {\url{https://universe.roboflow.com/491-resistor-classification/resistors_and_ceramic_caps}},
    url = {https://universe.roboflow.com/491-resistor-classification/resistors_and_ceramic_caps},
    journal = {Roboflow Universe},
    publisher = {Roboflow},
    year = {2026},
    month = {jan},
    note = {visited on 2026-03-22}
}

@misc{resistor_value_training_Computer_Vision_Model,
    title = {resistor value training Computer Vision Model},
    type = {Open Source Dataset},
    author = {ResistorV1},
    howpublished = {\url{https://universe.roboflow.com/resistorv1/resistor-value-training}},
    url = {https://universe.roboflow.com/resistorv1/resistor-value-training},
    journal = {Roboflow Universe},
    publisher = {Roboflow},
    note = {visited on 2026-03-22}
}

@article{chen2016resistor,
    title = {Computer vision on color-band resistor and its cost-effective diffuse light source design},
    author = {Chen, Yung-Sheng and Wang, Jeng-Yau},
    journal = {Journal of Electronic Imaging},
    volume = {25},
    number = {6},
    pages = {061409},
    year = {2016},
    publisher = {SPIE}
}

@inproceedings{reading2015resistor,
    author = {Yung-Sheng Chen and Jeng-Yau Wang},
    title = {Reading Resistor Based on Image Processing},
    booktitle = {2015 International Conference on Machine Learning and Cybernetics (ICMLC)},
    year = {2015},
    pages = {566--571},
    doi = {10.1109/ICMLC.2015.7340616}
}

@inproceedings{demir2018resistor,
    title = {Real-Time Resistor Color Code Recognition using Image Processing in Mobile Devices},
    author = {Demir, Muhammed Fatih and Cankirli, Aysenur and Karabatak, Begum and Yavariabdi, Amir and Mendi, Engin and Kusetogullari, Huseyin},
    booktitle = {2018 IEEE International Conference on Intelligent Systems (IS)},
    pages = {26--30},
    year = {2018},
    organization = {IEEE}
}

@article{serban2021resistor,
    title = {Computer vision algorithm for detecting resistor color codes},
    author = {Serban, N.-M. and Hobincu, R.},
    journal = {Romanian Journal of Information Science and Technology},
    volume = {24},
    number = {3},
    pages = {321--333},
    year = {2021}
}

@article{wibawanto2023resistor,
    title = {Ant Colony Optimization for Resistor Color Code Detection},
    author = {Wibawanto, Slamet and Kirana, Kartika Candra and Ramadhan, Hani},
    journal = {Knowledge Engineering and Data Science},
    volume = {6},
    number = {1},
    pages = {15--23},
    year = {2023},
    doi = {10.17977/um018v6i12023p15-23}
}

@article{liu2020resistor,
    title = {Method for Color-ring Resistor Detection and Localization
             in Printed Circuit Board Based on Convolutional Neural
             Network},
    author = {Liu, Xiaoyan and Li, Zhaoming and Duan, Jiaxu
              and Xiang, Tianyuan},
    journal = {Journal of Electronics \& Information Technology},
    volume = {42},
    number = {9},
    pages = {2302--2311},
    year = {2020},
    url = {https://jeit.ac.cn/en/article/doi/10.11999/JEIT190608},
    note = {Accessed: 2026-05-01}
}

@inproceedings{li2017resistor,
  author    = {Li, Xin and Zeng, Zi and Chen, Mei and Che, Shang-yue},
  title     = {A New Method of Resistor Color Ring Detection Based on Machine Vision},
  booktitle = {2017 Chinese Automation Congress (CAC)},
  year      = {2017},
  pages     = {241--245},
  doi       = {10.1109/CAC.2017.8242770}
}

@inproceedings{sokic2019resistor,
    title = {Automatic Segmentation and Classification of Resistors
             in Digital Images},
    author = {Muminovic, Mia and Sokic, Emir},
    booktitle = {2019 XXVII International Conference on Information,
                 Communication and Automation Technologies (ICAT)},
    pages = {1--6},
    year = {2019},
    organization = {IEEE},
    note = {{IEEE} Xplore document 8939034}
}

@misc{jocher2023ultralytics,
    author = {Jocher, Glenn and Chaurasia, Ayush and Qiu, Jing},
    title = {Ultralytics {YOLO} ({V}ersion 8.0.0)},
    year = {2023},
    howpublished = {\url{https://github.com/ultralytics/ultralytics}},
    note = {Accessed: 2026-04-27}
}

@inproceedings{zhou2018unetplusplus,
    title = {{UNet++}: A Nested {U-Net} Architecture for Medical Image Segmentation},
    author = {Zhou, Zongwei and Siddiquee, Md Mahfuzur Rahman and Tajbakhsh, Nima and Liang, Jianming},
    booktitle = {Deep Learning in Medical Image Analysis and Multimodal Learning for Clinical Decision Support (DLMIA)},
    pages = {3--11},
    year = {2018},
    publisher = {Springer}
}

@inproceedings{tan2019efficientnet,
    title = {{EfficientNet}: Rethinking Model Scaling for Convolutional Neural Networks},
    author = {Tan, Mingxing and Le, Quoc},
    booktitle = {International Conference on Machine Learning ({ICML})},
    pages = {6105--6114},
    year = {2019},
    organization = {PMLR}
}

@misc{SupreethRao99_CVResist,
    author = {Rao, Supreeth},
    title = {{CVResist}: Reads Resistor Values Using {OpenCV}},
    year = {2021},
    howpublished = {\url{https://github.com/SupreethRao99/CVResist}},
    note = {Accessed: 2026-04-27}
}

@article{aboyomi2023yolo,
    title = {A Comparative Analysis of Modern Object Detection Algorithms: {YOLO} vs. {SSD} vs. {Faster R-CNN}},
    author = {Aboyomi, Dalmar Dakari and Daniel, Cleo},
    journal = {ITEJ (Information Technology Engineering Journals)},
    volume = {8},
    number = {2},
    pages = {96--106},
    year = {2023},
    doi = {10.24235/itej.v8i2.123}
}

@misc{dpaml2024rc,
    author = {DPAML},
    title = {Rc Dataset},
    howpublished = {\url{https://universe.roboflow.com/dpaml-5bm7r/rc-zkutx}},
    url = {https://universe.roboflow.com/dpaml-5bm7r/rc-zkutx},
    year = {2024},
    publisher = {Roboflow},
    note = {Accessed: 2026-04-27}
}

@misc{iec60062,
    title = {{IEC} 60062:2016 --- {M}arking Codes for Resistors and Capacitors},
    author = {{International Electrotechnical Commission}},
    howpublished = {Edition 6.0},
    year = {2016},
    note = {{G}eneva, Switzerland}
}

@misc{iec60063,
    title = {{IEC} 60063:2015 --- {P}referred Number Series for Resistors and Capacitors},
    author = {{International Electrotechnical Commission}},
    howpublished = {Edition 3},
    year = {2015},
    note = {{G}eneva, Switzerland}
}

@inproceedings{lin2017focal,
    title = {{Focal Loss} for Dense Object Detection},
    author = {Lin, Tsung-Yi and Goyal, Priya and Girshick, Ross and He, Kaiming and Doll{\'a}r, Piotr},
    booktitle = {{IEEE} International Conference on Computer Vision ({ICCV})},
    pages = {2980--2988},
    year = {2017}
}

@inproceedings{milletari2016vnet,
    title = {{V-Net}: Fully Convolutional Neural Networks for Volumetric Medical Image Segmentation},
    author = {Milletari, Fausto and Navab, Nassir and Ahmadi, Seyed-Ahmad},
    booktitle = {2016 Fourth International Conference on 3D Vision (3DV)},
    pages = {565--571},
    year = {2016},
    organization = {IEEE}
}

@techreport{opus4.5,
  title         = {Claude {Opus} 4.5 System Card},
  author        = {Anthropic},
  year          = {2025},
  institution   = {Anthropic},
  url           = {https://www.anthropic.com/claude-opus-4-5-system-card}
}

@misc{GPT5.5,
  author       = {{OpenAI}},
  title        = {{GPT-5.5}},
  year         = {2026},
  url          = {https://openai.com},
}

@misc{gemini31pro,
  title = {Gemini 3.1 Pro},
  author = {{Google}},
  year = {2026},
  url = {https://deepmind.google/models/model-cards/gemini-3-1-pro/},
}

@misc{qwen2026qwen36,
  title        = {Qwen3.6: Large Language Model Series Update},
  author       = {{Qwen Team}},
  year         = {2026},
  month        = {April},
  url          = {https://qwen.ai/blog?id=qwen3.6},
}

\end{document}